\documentclass[12pt]{article}

\usepackage{amsmath,amsthm, amsfonts, amssymb, amsxtra, amsopn}
\usepackage{pgfplots}
\pgfplotsset{compat=1.13}
\usepackage{pgfplotstable}
\usepackage{graphicx,grffile}
\usepackage{multirow}
\usepackage{booktabs}
\usepackage{listings}
\usepackage{cmap}
\usepackage{colortbl}
\usepackage{adjustbox}
\usepackage{epsfig}
\usepackage{float}
\usepackage[british]{babel}
\usepackage{hhline}

\usepackage[tableposition=top]{caption}
\usepackage{subcaption}
\usepackage{makecell} 
\PassOptionsToPackage{hyphens}{url}

\usepackage{hyperref}
\hypersetup{colorlinks=true,linkcolor=black,citecolor=black,urlcolor=blue,filecolor=black}
\hypersetup{pdfpagemode=UseNone,pdfstartview=}

\usepackage{enumitem}
\setlist[itemize]{noitemsep, topsep=0pt}

\advance\oddsidemargin by -0.2in
\advance\textwidth by 0.4in

\advance\topmargin by -0.25in
\advance\textheight by 0.5in

\long\def\symbolfootnotetext[#1]#2{\begingroup%
\def\thefootnote{\fnsymbol{footnote}}\footnotetext[#1]{#2}\endgroup}

\def\zz{\phantom{0}}

\title{Predicting Pedestrian Crosswalk Behavior Using Convolutional Neural Networks}

\author{Eric Liang\footnotemark[1]\ \ \ 
Mark Stamp\footnotemark[1]\,\,\footnotemark[2]}

\begin{document}

\symbolfootnotetext[1]{Department of Computer Science, San Jose State University}
\symbolfootnotetext[2]{mark.stamp$@$sjsu.edu}

\maketitle

\abstract

A common yet potentially dangerous task is the act of crossing the street. Pedestrian accidents contribute a significant amount to the high number of annual traffic casualties, which is why it is crucial for pedestrians to use safety measures such as a crosswalk. However, people often forget to activate a crosswalk light or are unable to do so---such as those who are visually impaired or have occupied hands. Other pedestrians are simply careless and find the crosswalk signals a hassle, which can result in an accident where a car hits them. In this paper, we consider an improvement to the crosswalk system by designing a system that can detect pedestrians and triggering the crosswalk signal automatically. We collect a dataset of images that we then use to train a convolutional neural network to distinguish between pedestrians (including bicycle riders) and various false alarms. The resulting system can capture and evaluate images in real time, and the result can be used to automatically activate systems a crosswalk light. After extensive testing of our system in real-world environments, we conclude that it is feasible as a back-up system that can compliment existing crosswalk buttons, and thereby improve the overall safety of crossing the street.

\section{Introduction}

Traveling by foot or bicycle are popular transportation methods for people wanting to cover short distances. For such pedestrians, a common danger they face is car accidents when crossing a road. According to U.S. Centers for Disease Control and Prevention (CDC), there were more than~7,000 deaths and~104,000 emergency hospital visits for pedestrians in the year~2020, which clearly illustrates the need for better pedestrian safety~\cite{pedestrianAccidents}. There are already several safeguards for pedestrians crossing the road, with the most notable being the crosswalk. Activating a crosswalk increases the safety of pedestrians by over 30\% in some scenarios, thereby reducing the number of potential accidents~\cite{crosswalkEffectiveness}. However, there are still many cases where pedestrians do not use the crosswalk, whether from carelessness or an inability to do so. People with occupied hands may not be able to put their belongings down, and others who are visually impaired might have difficulty finding the button to activate the crosswalk lights. Furthermore, there are many people who simply ignore the crosswalk signals.

To reduce the number of accidents from pedestrians crossing the street, we consider a system that is complimentary to current crosswalk buttons. Specifically, we develop and test a system that can be used to automatically activate crosswalk lights for pedestrians in cases where the button is not pressed. We capture images and train a Convolutional Neural Network (CNN) to accomplish this task. CNNs, which loosely model the neurons in the human brain, can accurately classify previously unseen images~\cite{neuralNetwork}. CNNs were chosen for this research, since they achieve state of the art accuracy, and they are efficient enough to be used in a real-time, real-world environment.

To recognize pedestrians, we develop our own dataset containing images of the three primary objects that are in front of a crosswalk: pedestrians, bicyclers, and streets. Our goal is to ensure that the predictions of the program are sufficiently versatile and accurate for the purpose of automatically activating a crosswalk signal. We have also designed an experimental prototype system that incorporates a CNN to activate a crosswalk signal. To test our program, we deploy it at three different types of crosswalks and determine its accuracy. Our trained CNN achieves an accuracy in excess of~95\%, and when our system is deployed in a
%%%%% Not sure about the real-world case numbers--I think we can do better than this ?????
challenging real-world environment, it is able to automatically identify more than~85\% of pedestrians, 
with minimal false alarms.

The remainder of this paper is organized as follows. In Section~\ref{chap:dataset} we discuss the dataset 
collection process, as well as the preparation of the raw data for use in training and testing our CNN.
Section~\ref{chap:model} includes an outline of the specific CNN model that we employ, as well as
the hyperparameter tuning that we have employed when training the model on our image dataset.
Our experimental results are discussed in Section~\ref{chap:result}. Section~\ref{chap:future}
concludes the paper and provides suggestions for possible future work.

\section{Dataset}\label{chap:dataset}

In this section, we explain the process we used to collect a satisfactory dataset for our experiments. First, we consider the methodology behind the data collection, then we describe the processing needed to ensure that the dataset is compatible with our CNN.

\subsection{Data Collection Methodology}

One of the most important aspects of any machine learning model is the dataset used to train the model, as the quality of the model predictions are dependent on the quality of the training data. Consequently, it is important for us to collect data that accurately represents the classes that we want the model to distinguish. The dataset needs to be sufficiently varied to allow the model to generalize the classes so that it can accurately classify images that it has not previously seen. In practice, this means that our data needs to include a wide variety of subjects, as well as different angles, background settings, and so on.

For our crosswalk activation problem, we want to distinguish between the three most common views that can appear at a crosswalks: people walking (pedestrians), people biking (bikers), and an absence of people (street). Although pedestrians and bikers are both people who need to use the crosswalk button, we distinguish between the two classes in case users of the prediction system wish to trigger different actions based on a person's method of travel.

\begin{figure}[!htb]
    \centering
        \begin{tabular}{ccc}
        \includegraphics[width=0.25\textwidth]{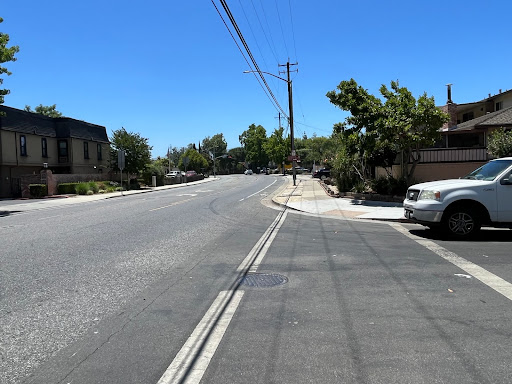} &
        \includegraphics[width=0.25\textwidth]{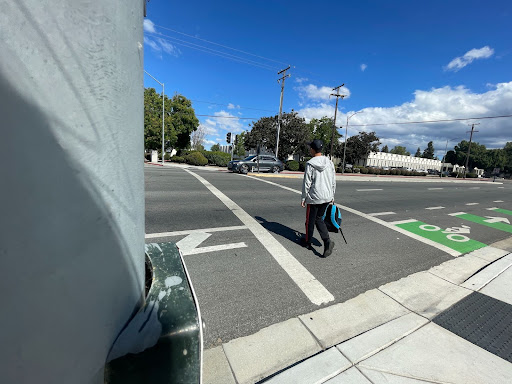} &
        \includegraphics[width=0.25\textwidth]{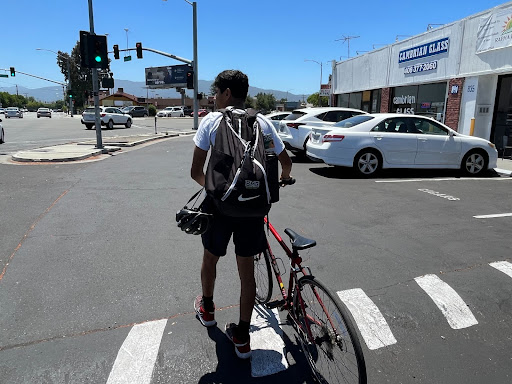} \\
        (a) Street &
        (b) Pedestrian &
        (c) Biker 
        \end{tabular}
    \caption{Sample images from our dataset}\label{fig:class_sample}
\end{figure}

We have manually collected pictures for our dataset, as this allows us the greatest flexibility with respect to angles, locations, and other important factors of the images. At this point, our system is experimental, so there is no need to use automatic cameras to produce the data. A sample images of each class type appears in Figure~\ref{fig:class_sample}.

In addition to the contents of the images, it is also important to ensure that the relative sizes of the classes do not contribute any bias. One common form of bias occurs from unbalanced numbers of samples, that is, when one class has far more samples than another class. This can result in a model that is more tuned for the over-represented class, resulting in more predictions for that class, which can yield a higher accuracy, but a lower balanced accuracy~\cite{StampMLbook}. Therefore, we have constructed our dataset with the ratios between classes similar to what we might expect to see in practice. Our dataset includes a total of~2000 images, of which~207 are bikers, 668 are pedestrians, and~1125 are streets. The ratio of positive to negative training images is 7:9, which does not over-represent either the negative class (street) or positive classes (pedestrians and bikers). We also have a ratio of roughly~1:3.2 between bikers and pedestrians, so as to account for the greater number of pedestrians that we expect to see crossing the street, as compared to bike riders.

\subsection{Compatible Dataset}

\begin{figure}[!htb]
    \centering
        \begin{tabular}{ccc}
        \includegraphics[width=0.25\textwidth]{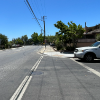} &
        \includegraphics[width=0.25\textwidth]{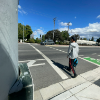} & 
        \includegraphics[width=0.25\textwidth]{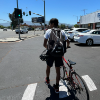} \\
        (a) Street &
        (b) Pedestrian &
        (c) Biker 
        \end{tabular}
    \caption{Down-scaled and cropped versions of dataset images}\label{fig:class_sample_pixel}
\end{figure}

The original resolution of our images is~$4032\times 3024$, which is far too large for training a CNN. Although higher resolution images contain more detail, the training time for such large images would be prohibitive. Therefore, we have 
down-sampled all of our images to~$100\times 100$ pixels, which is sufficient to maintain recognizable detail, 
while improving training and classification efficiency. The down-sampled versions of the images in Figure~\ref{fig:class_sample} appear in Figure~\ref{fig:class_sample_pixel}.

In order to transform the images into a dataset that is usable by our CNN model, each image has to be loaded as a tensor data type and represented as a class. We use the software tools listed in Table~\ref{dataset_tools} to prepare our images for CNN training and classification.

\begin{table}[!htb]
    \caption{Tools used to load dataset\label{dataset_tools}}
    \centering
    \adjustbox{scale=0.85}{
    \begin{tabular}{c|l}\midrule\midrule
            Tool & Purpose\\
    \midrule %%%%% Cite a reference for each of these tool ?????
            \texttt{PyTorch} & Provides the necessary ``tensor'' data type ~\cite{pyTorch}\\
            \texttt{Dataset} & Creates a class to communicate with CNN model ~\cite{pyTorch} \\
            \texttt{pandas} & Reads the images into the program ~\cite{pandas}\\
            \texttt{sci-kit image} & Converts images into tensor ~\cite{scikit-image}\\
    \midrule\midrule
    \end{tabular}
    }
\end{table}

\section{Machine Learning Model}\label{chap:model}

In this section, we first discuss the Convolutional Neural Network (CNN) model that we have chosen in detail, including the process and tools used to create the model. Then we consider the various hyperparameters and tune these parameters with the goal of optimizing the accuracy of our CNN.

\subsection{CNN Model}

CNNs are a widely used class of deep learning models that were developed for classifying images. Similar to other neural networks, CNNs contain multiple layers and assigns weights to different parts of the input. However, a unique feature of the CNN is its convolutional layers which, in effect, scan the image and model the important features of an image. Through multiple convolutional layers, the CNN is able to achieve higher levels of abstraction, and ultimately classify images via a fully connected output layer~\cite{convNet}. Due to their convolutional layers, CNNs provide a high degree of translation invariance, which is crucial when dealing with images. As a result, CNNs avoid the overfitting that tends to occur when fully connected models are used with images, while also vastly reducing the number of weights that must be learned, thereby offering much greater training efficiency.

For our CNN model, we employ transfer learning, that is, we employ a model that has been pre-trained on a vast image dataset, and only retrain the output layer. In this way, we can take advantage of the learning that is represented by the hidden layers of the pre-trained model, while training the model for the task at hand, namely, distinguishing between images of pedestrians, bikers, and streets. Specifically, we use the GoogLeNet convolutional neural network, which is a pre-trained architecture with~22 layers. Using a pre-trained model allows us to save a great deal of training time without sacrificing accuracy. The GoogLeNet architecture is a widely used model for computer vision that has proven to be both fast and accurate, and is suitable for real-time models~\cite{googleNet}. The various layers and interconnections of the GoogLeNet architecture are shown in Figure~\ref{fig:googlenet} in the Appendix. 

Due to its efficiency and accuracy, we use the Adam optimizer when training our model. The Adam optimizer is a very efficient first-gradient optimizer that has a limited memory requirement, making it ideal for training our pedestrian prediction system~\cite{adam}.

%%%%% Need a reference for this. Also, might be a bit TMI (too much information) ?????
The implementation of the GoogLeNet CNN model is relatively simple, as many of the tools already exist in \texttt{PyTorch} ~\cite{pyTorch}. During each iteration of training the model, we loop through the entire batch and calculate the loss of each item by testing it with the current model. We use the built-in cross entropy loss, as it is useful for classification problems ~\cite{pyTorch}. Afterwards, the algorithm  ``learns'' from the batch and changes the model based on the Adam optimizer. After many iterations, the model is done training and can evaluate images that are fed into it.

\subsection{Hyperparameter Tuning}

Hyperparameters are model parameters that must be set by a user before training a model. Tuning hyperparameters is
a vital aspect of training any machine learning model, as the hyperparameters can have a major impact on the resulting accuracy~\cite{hyperparameters}. Since certain hyperparameters are already defined in the GoogLeNet architecture, we only have to optimize the learning rate, number of epochs, and the batch size of our model. The learning rate is, in effect, the step size during gradient descent, and it can affect both the speed and accuracy of training. Although smaller learning rates allow for more reliable convergence in gradient descent, it also takes longer to converge~\cite{hyperparameters}. The batch size determines the number of the training sample given to the model considers at each iteration. The number of epochs determines the number of times the entire training dataset is considered during training, affecting both speed and accuracy, as well as issues related to overfitting and underfitting. It is important to tune these hyperparameters to optimize the accuracy, efficiency, and to avoid common problems, such as overfitting~\cite{hyperparameters}.

We identify suitable hyperparameter values by using a grid search over the learning rate, batch size, and number of epochs. This allows us to determine a combination of these three values that is optimal within our search space. When testing the hyperparameters, we use~1900 images for training the model and~100 images for testing the model. The hyperparameter values tested are listed in Table~\ref{tab:parameters}, with the selected optimal values given in boldface.

\begin{table}[!htb]
\centering
\caption{Hyperparameters tested and selected}\label{tab:parameters}
        \adjustbox{scale=0.9}{
        \begin{tabular}{c|cc|c}\midrule\midrule
            \multirow{2}{*}{Parameters} & 
            \multirow{2}{*}{Description} & 
            \multirow{2}{*}{Tested values} & Average\\
                                &                    &                        & accuracy \\
        \midrule
            \texttt{learning rate} & Step size & 0.0001, \textbf{0.0005}, 0.001, 0.01 \\ 
            \texttt{batch size} & Images per iteration & 8, 16, \textbf{32}, 64 & 0.9567\\
            \texttt{epochs} & Passes through data & 2, \textbf{4}, 6, 8\\
        \midrule\midrule
        \end{tabular}
        }
\end{table}

\begin{figure}[!htb]
    \centering
        \begin{tabular}{cc}
        \includegraphics[width=0.45\textwidth]{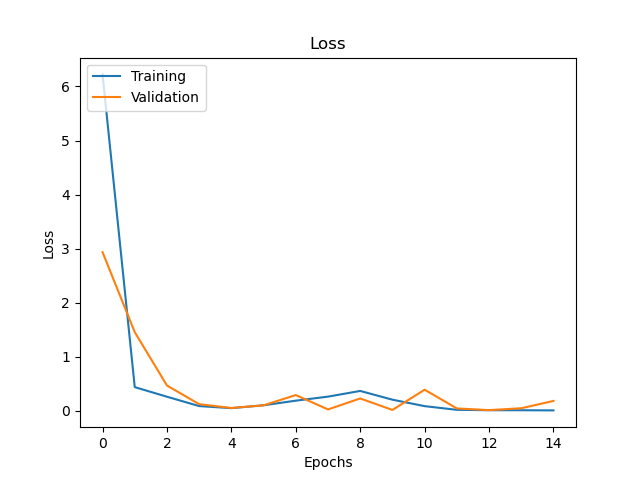} &

        \includegraphics[width=0.45\textwidth]{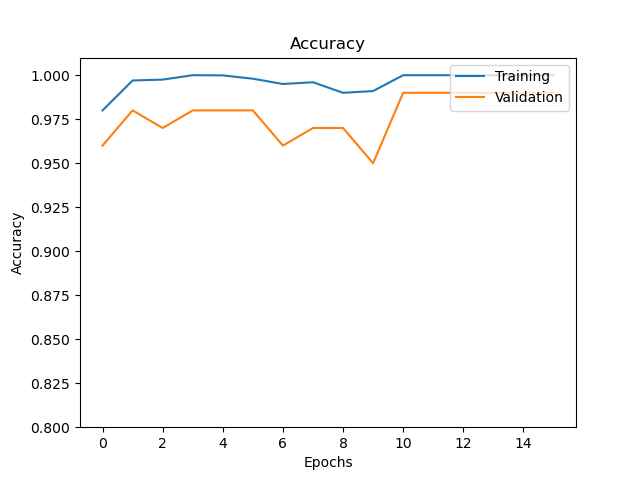} \\
        (a) Loss &
        (b) Accuracy
        \end{tabular}
    \caption{CNN Model loss and accuracy}\label{fig:accuracy_loss_graphs}
\end{figure}

The results of our model training is summarized in Figure~\ref{fig:accuracy_loss_graphs}, depicting the accuracy and loss graphs for both training and validation phases of our GoogLeNet model. The training and validation accuracy track closely, as do the loss plots, indicating that overfitting is not a significant issue for the models when trained through the~14 epochs considered.
We also observe that near-optimal accuracy and loss are achieved with~4 epochs, so for the sake of efficiency,
we train for~4 epochs in all subsequent experiments.

Note that using the optimal hyperparameters indicated in Table~\ref{tab:parameters}, 
the overall accuracy for our three classes classification problem (i.e., pedestrians, bikers, and streets)
is~0.9567. Furthermore, the model is relatively quick to train and validate, 
requiring roughly~850 seconds for~4 epochs. 

\section{Experiments and Results}\label{chap:result}

In this section, we describe our experiments and analyze our results. We first outline the software created 
to utilize our CNN model in various real-time scenarios, and we present results for these different scenarios.

\subsection{Prediction System}

\begin{figure}[!htb]
    \centering
    \includegraphics[width=0.825\textwidth]{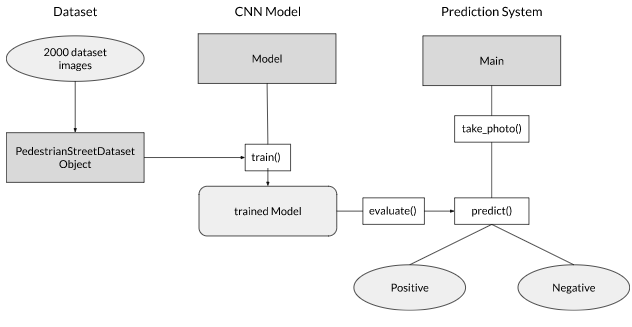}
    \caption{The process from training the model to predicting images}
    \label{fig:prediction_lifecycle}
\end{figure}

Based on the results from Section~\ref{chap:model}, our CNN model has an accuracy in excess of~95\%.
Our goal here is to deploy this CNN in a real-time experimental pedestrian prediction system. 
We do so by creating a program that captures images with a short delay of~1 second and evaluates 
the images using the prediction CNN discussed in the previous section. By keeping track of and comparing the current result with previous results, this program output can then be used to trigger
actions (e.g., crosswalk lights) if a pedestrian or biker wanting to cross the street is detected. 
Our software uses beeps to signify that a pedestrian or biker has been detected, but these
can be mapped to traffic control systems such as crosswalk signals. 
A high-level view of the relationship between the CNN and 
the prediction system is given in Figure~\ref{fig:prediction_lifecycle}.

\subsection{Thresholding}\label{sect:imageCap}

To reduce false positives that would needlessly activate the crosswalk signal, 
we consider a series of images and a threshold for the number of 
positive images, where a ``positive'' image is one that is classified by the CNN
as a pedestrian or biker. The~$n$ most recent images are considered, 
for each value of~$n\in\{1,2,3,4,5,6,7\}$, and all thresholds~$t=0,1,\ldots,n$,
where a threshold of~$t=0$ is defined as a pedestrian or biker always being identified.

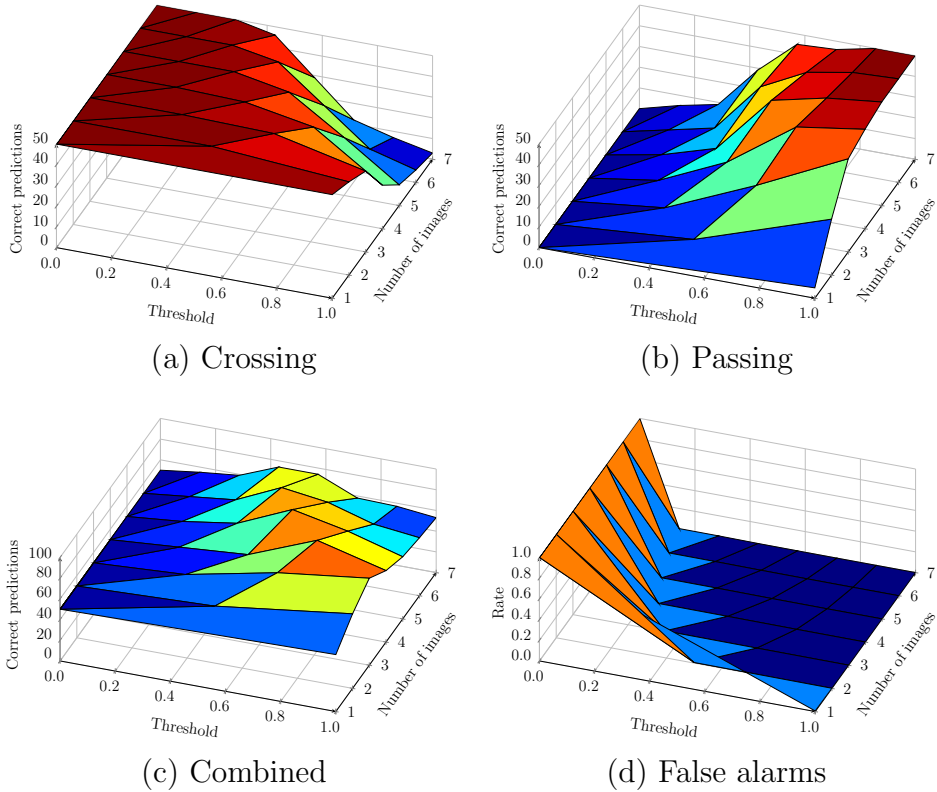
\begin{figure}[!htb]
    \centering\setlength{\tabcolsep}{2pt}
    \begin{tabular}{cc}
    \adjustbox{scale=0.5}{
     \begin{tikzpicture}[xscale=1.0,yscale=1.0]
%  \begin{axis} 
%   \addplot3[surf] coordinates
\begin{axis}[%
%view={45}{110},
%view={64}{26},
%view={10}{26},
y dir = reverse,
%view={10}{40},
view={-70}{55},
width=10cm,
height=7.8cm,
scale only axis,
xmin=1, xmax=7,
xmajorgrids,
ymin=0, 
ymax=1,
ymajorgrids,
zmin=0, zmax=50,
zmajorgrids,
axis lines=left,
grid=major,
%xlabel=Record Length,
xlabel=Number of images,
xlabel style={xshift=0.2cm,yshift=0.2cm,rotate=55},
ylabel=Threshold,
ylabel style={rotate=-8},
xtick = {1,2,3,4,5,6,7},
ytick = {0.00,0.20,0.40,0.60,0.80,1.00},
ztick = {0,10,20,30,40,50},
zlabel=Correct predictions,
zlabel style={xshift=0.6cm}, %,yshift=-0.1cm},
xticklabel style={xshift=0.1cm, yshift=0.3cm,
    		/pgf/number format/.cd,
		1000 sep={},
    		fixed,
    		fixed zerofill,
    		precision=0},
yticklabel style={%xshift=0.15cm, yshift=0.45cm,
    		/pgf/number format/.cd,
		1000 sep={},
    		fixed,
    		fixed zerofill,
    		precision=1},
zticklabel style={yshift=0.3cm,
    		/pgf/number format/.cd,
		1000 sep={},
    		fixed,
    		fixed zerofill,
    		precision=0}]
%z filter/.code={\pgfmathparse{#1/100}\pgfmathresult}]
%
\addplot3[%
surf,
z buffer=sort,
colormap/jet,
shader=flat,
draw=black]
coordinates{ 

(1,0,50)
(1,0,50)
(1,0,50)
(1,0,50)
(1,0,50)
(1,0,50)
(1,0,50)
(1,1,50)

(2,0,50)
(2,0,50)
(2,0,50)
(2,0,50)
(2,0,50)
(2,0,50)
(2,0.5,50)
(2,1,45)

(3,0,50)
(3,0,50)
(3,0,50)
(3,0,50)
(3,0,50)
(3,0.3333,50)
(3,0.6667,49)
(3,1,39)

(4,0,50)
(4,0,50)
(4,0,50)
(4,0,50)
(4,0.25,50)
(4,0.5,49)
(4,0.75,43)
(4,1,21)

(5,0,50)
(5,0,50)
(5,0,50)
(5,0.2,50)
(5,0.4,50)
(5,0.6,45)
(5,0.8,28)
(5,1,10)

(6,0,50)
(6,0,50)
(6,0.1667,50)
(6,0.3333,50)
(6,0.5,46)
(6,0.6667,29)
(6,0.8333,12)
(6,1,6)

(7,0,50)
(7,0.1429,50)
(7,0.2857,50)
(7,0.4286,46)
(7,0.5714,29)
(7,0.7143,12)
(7,0.8571,7)
(7,1,3)

};
  \end{axis}
  \end{tikzpicture}
    }
    &
    \adjustbox{scale=0.5}{
     \begin{tikzpicture}[xscale=1.0,yscale=1.0]
%  \begin{axis} 
%   \addplot3[surf] coordinates
\begin{axis}[%
%view={45}{110},
%view={64}{26},
%view={10}{26},
y dir = reverse,
%view={10}{40},
view={-70}{55},
width=10cm,
height=7.8cm,
scale only axis,
xmin=1, xmax=7,
xmajorgrids,
ymin=0, 
ymax=1,
ymajorgrids,
zmin=0, zmax=50,
zmajorgrids,
axis lines=left,
grid=major,
%xlabel=Record Length,
xlabel=Number of images,
xlabel style={xshift=0.2cm,yshift=0.2cm,rotate=55},
ylabel=Threshold,
ylabel style={rotate=-8},
xtick = {1,2,3,4,5,6,7},
ytick = {0.00,0.20,0.40,0.60,0.80,1.00},
ztick = {0,10,20,30,40,50},
zlabel=Correct predictions,
zlabel style={xshift=0.6cm}, %,yshift=-0.1cm},
xticklabel style={xshift=0.1cm, yshift=0.3cm,
    		/pgf/number format/.cd,
		1000 sep={},
    		fixed,
    		fixed zerofill,
    		precision=0},
yticklabel style={%xshift=0.15cm, yshift=0.45cm,
    		/pgf/number format/.cd,
		1000 sep={},
    		fixed,
    		fixed zerofill,
    		precision=1},
zticklabel style={yshift=0.3cm,
    		/pgf/number format/.cd,
		1000 sep={},
    		fixed,
    		fixed zerofill,
    		precision=0}]
%z filter/.code={\pgfmathparse{#1/100}\pgfmathresult}]
%
\addplot3[%
surf,
z buffer=sort,
colormap/jet,
shader=flat,
draw=black]
coordinates{ 

(1,0,0)
(1,0,0)
(1,0,0)
(1,0,0)
(1,0,0)
(1,0,0)
(1,0,0)
(1,1,5)

(2,0,0)
(2,0,0)
(2,0,0)
(2,0,0)
(2,0,0)
(2,0,0)
(2,0.5,5)
(2,1,27)

(3,0,0)
(3,0,0)
(3,0,0)
(3,0,0)
(3,0,0)
(3,0.3333,5)
(3,0.6667,24)
(3,1,45)

(4,0,0)
(4,0,0)
(4,0,0)
(4,0,0)
(4,0.25,5)
(4,0.5,20)
(4,0.75,43)
(4,1,48)

(5,0,0)
(5,0,0)
(5,0,0)
(5,0.2,5)
(5,0.4,13)
(5,0.6,41)
(5,0.8,47)
(5,1,50)

(6,0,0)
(6,0,0)
(6,0.1667,5)
(6,0.3333,10)
(6,0.5,34)
(6,0.6667,45)
(6,0.8333,49)
(6,1,50)

(7,0,0)
(7,0.1429,5)
(7,0.2857,9)
(7,0.4286,29)
(7,0.5714,45)
(7,0.7143,46)
(7,0.8571,50)
(7,1,50)

};
  \end{axis}
  \end{tikzpicture}
    }
    \\
    (a) Crossing
    &
    (b) Passing
    \\ \\
    \adjustbox{scale=0.5}{
     \begin{tikzpicture}[xscale=1.0,yscale=1.0]
%  \begin{axis} 
%   \addplot3[surf] coordinates
\begin{axis}[%
%view={45}{110},
%view={64}{26},
%view={10}{26},
%view={-70}{50},
%view={10}{40},
y dir = reverse,
%view={30}{40},
view={-70}{55},
width=10cm,
height=7.8cm,
scale only axis,
xmin=1, xmax=7,
xmajorgrids,
ymin=0, 
ymax=1,
ymajorgrids,
zmin=0, zmax=100,
zmajorgrids,
axis lines=left,
grid=major,
%xlabel=Record Length,
xlabel=Number of images,
xlabel style={xshift=0.2cm,yshift=0.2cm,rotate=55},
ylabel=Threshold,
ylabel style={rotate=-8},
xtick = {1,2,3,4,5,6,7},
ytick = {0.00,0.20,0.40,0.60,0.80,1.00},
ztick = {0,20,40,60,80,100},
zlabel=Correct predictions,
zlabel style={xshift=0.6cm}, %,yshift=-0.1cm},
xticklabel style={xshift=0.1cm, yshift=0.3cm,
    		/pgf/number format/.cd,
		1000 sep={},
    		fixed,
    		fixed zerofill,
    		precision=0},
yticklabel style={%xshift=0.15cm, yshift=0.45cm,
    		/pgf/number format/.cd,
		1000 sep={},
    		fixed,
    		fixed zerofill,
    		precision=1},
zticklabel style={yshift=0.3cm,
    		/pgf/number format/.cd,
		1000 sep={},
    		fixed,
    		fixed zerofill,
    		precision=0}]
%z filter/.code={\pgfmathparse{#1/100}\pgfmathresult}]
%
\addplot3[%
surf,
z buffer=sort,
colormap/jet,
shader=flat,
draw=black]
coordinates{ 

(1,0,50)
(1,0,50)
(1,0,50)
(1,0,50)
(1,0,50)
(1,0,50)
(1,0,50)
(1,1,55)

(2,0,50)
(2,0,50)
(2,0,50)
(2,0,50)
(2,0,50)
(2,0,50)
(2,0.5,55)
(2,1,72)

(3,0,50)
(3,0,50)
(3,0,50)
(3,0,50)
(3,0,50)
(3,0.3333,55)
(3,0.6667,73)
(3,1,84)

(4,0,50)
(4,0,50)
(4,0,50)
(4,0,50)
(4,0.25,55)
(4,0.5,60)
(4,0.75,86)
(4,1,69)

(5,0,50)
(5,0,50)
(5,0,50)
(5,0.2,55)
(5,0.4,63)
(5,0.6,86)
(5,0.8,75)
(5,1,60)

(6,0,50)
(6,0,50)
(6,0.1667,55)
(6,0.3333,60)
(6,0.5,80)
(6,0.6667,74)
(6,0.8333,61)
(6,1,56)

(7,0,50)
(7,0.1429,55)
(7,0.2857,59)
(7,0.4286,74)
(7,0.5714,74)
(7,0.7143,58)
(7,0.8571,57)
(7,1,53)

};
  \end{axis}
  \end{tikzpicture}
    }
    &
    \adjustbox{scale=0.5}{
     \begin{tikzpicture}[xscale=1.0,yscale=1.0]
%  \begin{axis} 
%   \addplot3[surf] coordinates
\begin{axis}[%
%view={45}{110},
%view={64}{26},
%view={10}{26},
y dir = reverse,
%view={10}{40},
view={-70}{55},
width=10cm,
height=7.8cm,
scale only axis,
xmin=1, xmax=7,
xmajorgrids,
ymin=0, 
ymax=1,
ymajorgrids,
zmin=0, zmax=1,
zmajorgrids,
axis lines=left,
grid=major,
%xlabel=Record Length,
xlabel=Number of images,
xlabel style={xshift=0.2cm,yshift=0.2cm,rotate=55},
ylabel=Threshold,
ylabel style={rotate=-8},
xtick = {1,2,3,4,5,6,7},
ytick = {0.00,0.20,0.40,0.60,0.80,1.00},
ztick = {0.00,0.20,0.40,0.60,0.80,1.00},
zlabel=Rate,
zlabel style={xshift=0.6cm}, %,yshift=-0.1cm},
xticklabel style={xshift=0.1cm, yshift=0.3cm,
    		/pgf/number format/.cd,
		1000 sep={},
    		fixed,
    		fixed zerofill,
    		precision=0},
yticklabel style={%xshift=0.15cm, yshift=0.45cm,
    		/pgf/number format/.cd,
		1000 sep={},
    		fixed,
    		fixed zerofill,
    		precision=1},
zticklabel style={yshift=0.3cm,
    		/pgf/number format/.cd,
		1000 sep={},
    		fixed,
    		fixed zerofill,
    		precision=1}]
%z filter/.code={\pgfmathparse{#1/100}\pgfmathresult}]
%
\addplot3[%
surf,
z buffer=sort,
colormap/jet,
shader=flat,
draw=black]
coordinates{ 

(1,0,1)
(1,0,1)
(1,0,1)
(1,0,1)
(1,0,1)
(1,0,1)
(1,0,1)
(1,1,0.00595)

(2,0,1)
(2,0,1)
(2,0,1)
(2,0,1)
(2,0,1)
(2,0,1)
(2,0.5,0.00595)
(2,1,0.00119)

(3,0,1)
(3,0,1)
(3,0,1)
(3,0,1)
(3,0,1)
(3,0.3333,0.00595)
(3,0.6667,0.00298)
(3,1,0.00060)

(4,0,1)
(4,0,1)
(4,0,1)
(4,0,1)
(4,0.25,0.00595)
(4,0.5,0.00417)
(4,0.75,0.00119)
(4,1,0)

(5,0,1)
(5,0,1)
(5,0,1)
(5,0.2,0.00595)
(5,0.4,0.00476)
(5,0.6,0.00119)
(5,0.8,0)
(5,1,0)

(6,0,1)
(6,0,1)
(6,0.1667,0.00595)
(6,0.3333,0.00476)
(6,0.5,0.00179)
(6,0.6667,0.00060)
(6,0.8333,0)
(6,1,0)

(7,0,1)
(7,0.1429,0.00595)
(7,0.2857,0.00536)
(7,0.4286,0.00179)
(7,0.5714,0.00060)
(7,0.7143,0)
(7,0.8571,0)
(7,1,0)

};
  \end{axis}
  \end{tikzpicture}
    }
    \\
    (c) Combined
    &
%    (d) False positives
    (d) False alarms
    \end{tabular}
    \caption{Thresholding results} 
    \label{fig:aaa_confidence}
\end{figure}

For each number of consecutive images~$n$, 
we test~50 instances of a pedestrian (or biker) walking past the camera without stopping and~50 instances 
of a pedestrian (or biker) stopping in front of the camera for a specified amount of time. 
We vary the time it takes for people to walk past the camera or wait in front of the camera to cross the street. 
A prediction is correct if the model does not predict ``pedestrian'' or ``biker'' when a person is passing by,
or it predicts ``pedestrian'' or ``biker'' at least once before 
a person crosses the street; otherwise, the prediction is incorrect. 
The results of these experiments are summarized in the graphs in 
Figure~\ref{fig:aaa_confidence}. 

%\begin{figure}[!htb]
%    \centering
%    \input figures/aaa_crossing_confidence_factor.tex
%    \caption{Confidence factor (crossing)}
%    \label{fig:aaa_crossing_confidence_factor}
%\end{figure}

%\begin{figure}[!htb]
%    \centering
%    \input figures/aaa_passing_confidence_factor.tex
%    \caption{Confidence factor (passing)}
%    \label{fig:aaa_passing_confidence_factor}
%\end{figure}

%\begin{figure}[!htb]
%    \centering
%    \input figures/aaa_total_confidence_factor.tex
%    \caption{Confidence factor (total)} 
%    \label{fig:aaa_total_confidence_factor}
%\end{figure}

%\begin{figure}[!htb]
%    \centering
%    \input figures/aaa_false positives.tex
%    \caption{Confidence factor (false positives)} 
%    \label{fig:aaa_false-positives}
%\end{figure}

Note that Figure~\ref{fig:aaa_confidence}(c) gives
the accuracies, based on the combined results from Figures~\ref{fig:aaa_confidence}(a) 
and~\ref{fig:aaa_confidence}(b). Also, Figures~\ref{fig:aaa_confidence}(d) gives the
false alarm rate, where a false alarm is defined as a pedestrian or biker 
being detected when neither is present. That is, false alarms occur
when images of the street---as opposed to images of
someone intending to cross the street or someone passing by---result in
a misclassification that would trigger the crosswalk light.
Note also that for a threshold of~0, we assume that 
a classification of either pedestrian or biker always results,
which implies that the crosswalk signal would always be activated.
This arbitrary assumption makes the graphs in Figure~\ref{fig:aaa_confidence}
clearer.

In terms of accuracy, the best results are summarized in Table~\ref{tab:record}.
We observe that~$n=4$ images with a threshold of~$t=3$ balances the
accuracy for detecting passing and crossing people, while the~$(n,t)=(3,3)$
case is slightly better at identifying passing people, while the~$(n,t)=(5,3)$
case is slightly better at identifying crossings. Thus, the optimal choice of~$(n,t)$
would depend on whether a crosswalk tends to have more people passing
by or crossing the street. Also, smaller~$n$ implies a shorter wait time, 
which could be a factor to consider. For all subsequent experiments,
we have used~$(n,t)=(5,3)$.

%\begin{table}[!htb]
%\centering
%\caption{Record lengths tested and selected}\label{tab:record}
%        \adjustbox{scale=0.825}{
%        \begin{tabular}{c|c|c|c|c}\midrule\midrule
%            Record & Passing people & Crossing people & Total people & Combined\\
%            length & correctly identified & correctly identified & correctly identified & accuracy\\
%        \midrule %%%%% Add a column for "combined accuracy" ?????
%            3 & 28/50 & 49/50 & 77/100 & 0.77 \\ 
%            \textbf{5} & 42/50 & 45/50 & 87/100 & 0.87 \\ 
%            7 & 46/50 & 36/50 & 82/100 & 0.82 \\ 
%            9 & 50/50 & 29/50 & 79/100 & 0.79 \\ 
%        \midrule\midrule
%        \end{tabular}
%        }
%\end{table}

\begin{table}[!htb]
\centering
\caption{Best accuracy results}\label{tab:record}
        \adjustbox{scale=0.8}{
        \begin{tabular}{c|c|c|c|c|c}\midrule\midrule
            Number & \multirow{2}{*}{Threshold} & Passing people     & Crossing people    & \multirow{2}{*}{Accuracy} & False\\
            of images &                                        & correctly identified & correctly identified & & positives\\
        \midrule %%%%% Add a column for "combined accuracy" ?????
            3 & 3 & 45/50 & 39/50 & 0.84 & 1\\ 
            4 & 3 & 43/50 & 43/50 & 0.86 & 2\\ 
            5 & 3 & 41/50 & 45/50 & 0.86 & 2\\ 
        \midrule\midrule
        \end{tabular}
        }
\end{table}

%Based on the results in Table~\ref{tab:record}, we consider a record of~5 predictions to be best,
%as it yields the highest combined accuracy. When the record length is too short, a pedestrian or biker
%only needs to be in the program view for a few seconds to trigger the crosswalk even when a person is only passing by. 
%On the other hand, when the record length is too long, the program fails to detect some of the people 
%waiting in front of the camera because they begin to cross before positive predictions reach a majority. 
%A record length of~5 can still fail, if a person passes slowly without crossing, 
%or a person crosses after an unusually short waiting time.

\subsection{Real-World Performance}

To better analyze the accuracy of our pedestrian prediction system, 
we deployed our model in various real-world environments. While the experiments 
in Section~\ref{sect:imageCap} consisted of test subjects, the experiments reported
in this section are from actual users at three distinct crosswalk locations. We determine the predictions 
that the CNN model makes and compare them with a recorded video of the camera's view,
from which we determine whether a subject actually crossed the street or 
simply passed by without crossing the street.

To account for the different settings that a pedestrian prediction system might be used in, 
we have conducted this experiment at three different locations. These locations are the following.
\begin{enumerate}
\item A crosswalk next to a summer school at the end of the school day
\item A busy traffic intersection next to a shopping mall
\item A crosswalk next to a park
\end{enumerate}
These locations were also chosen as they represent streets where implementing our automated
crosswalk safety system would likely be beneficial. Each experiment consists of a one hour time interval, 
or~3600 predictions.
For each crosswalk, we tally the total number of unique pedestrians and bikers that are both passing by and waiting to 
cross the street. Based on the recording of the program beeps, we determine whether the pedestrian 
prediction model correctly identified the person's street crossing (or not) behavior. These metrics are also 
broken down separately for pedestrians and bikers. Finally, we identify the number of false alarms 
that occur during the one hour span, where a pedestrian was predicted when the model should identify 
a street. These results are presented in Table~\ref{tab:predictions} and Table~\ref{tab:false_positives}.

\begin{table}[!htb]
\caption{Prediction results at different locations}\label{tab:predictions}
    \centering
    %%%%% We need rates listed in this table ?????
    \begin{tabular}{c}
    (a) Location 1 (summer school crosswalk) \\
        \adjustbox{scale=0.775}{
          \begin{tabular}{p{5cm}|c|c|c|c}
            \midrule\midrule
            \multirow{2}{5cm}{Method of transportation} 
            & \multicolumn{2}{c|}{No intent to cross (passing)} 
            & \multicolumn{2}{c}{Intent to cross (waiting)}\\
             \cmidrule{2-5}
            & Correctly identified & Total & Correctly identified & Total\\ \midrule
            Pedestrians & 13  & 16 & 74 & 79 \\ 
            Bike riders & \zz7 & \zz7 & \zz9 & 12 \\ \midrule
            Total & 20 & 23 & 83 & 91 \\ \midrule
            Accuracy
            & \multicolumn{2}{c|}{0.8700}
            & \multicolumn{2}{c}{0.9121}\\ \midrule
            Combined accuracy
            & \multicolumn{4}{c}{0.9035}\\ \midrule\midrule
          \end{tabular}
        } \\ \\
    (b) Location 2 (shopping mall traffic intersection) \\
        \adjustbox{scale=0.775}{
          \begin{tabular}{p{5cm}|c|c|c|c}
            \midrule\midrule
            \multirow{2}{5cm}{Method of transportation} 
            & \multicolumn{2}{c|}{No intent to cross (passing)} 
            & \multicolumn{2}{c}{Intent to cross (waiting)}\\
             \cmidrule{2-5}
            & Correctly identified & Total & Correctly identified & Total\\ \midrule
            Pedestrians & 63  & 79 & 115 & 128 \\ 
            Bike riders & \zz6 & \zz7 & \zz10 & \zz15 \\ \midrule
            Total & 69 & 86 & 125 & 143 \\ \midrule
            Accuracy
            & \multicolumn{2}{c|}{0.8023}
            & \multicolumn{2}{c}{0.8741}\\ \midrule
            Combined accuracy
            & \multicolumn{4}{c}{0.8472}\\ \midrule\midrule
          \end{tabular}
        } \\ \\
    (c) Location 3 (park crosswalk) \\
        \adjustbox{scale=0.775}{
          \begin{tabular}{p{5cm}|c|c|c|c}
            \midrule\midrule
            \multirow{2}{5cm}{Method of transportation} 
            & \multicolumn{2}{c|}{No intent to cross (passing)} 
            & \multicolumn{2}{c}{Intent to cross (waiting)}\\
             \cmidrule{2-5}
            & Correctly identified & Total & Correctly identified & Total\\ \midrule
            Pedestrians & 36  & 51 & 32 & 35 \\ 
            Bike riders & 14 & 15 & \zz5 & \zz8 \\ \midrule
            Total & 50 & 66 & 37 & 43 \\ \midrule
            Accuracy
            & \multicolumn{2}{c|}{0.7576}
            & \multicolumn{2}{c}{0.8605}\\ \midrule
            Combined accuracy
            & \multicolumn{4}{c}{0.7982}\\ \midrule\midrule
          \end{tabular}
        }
\end{tabular}
\end{table}

\begin{table}[!htb]
\centering
%%%%% Also need to include rates here ?????
\caption{False alarms per location}\label{tab:false_positives}
        \adjustbox{scale=0.9}{
        \begin{tabular}{c|c|c|c}\midrule\midrule
            Location & Pedestrians & Bike riders & False alarm rate\\
        \midrule
            Summer school & 0 & 0 & 0.00000\\ 
            Shopping mall & 1 & 0 & 0.00028 \\ 
            Park & 2 & 1 & 0.00083\\ \midrule
            Average & 1.00 & 0.33 & 0.00037\\
        \midrule\midrule
        \end{tabular}
        }
\end{table}

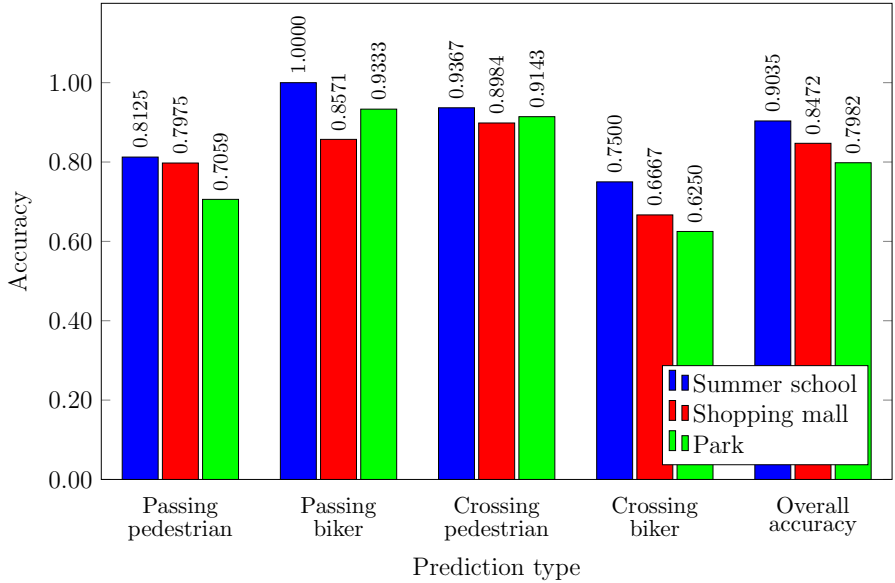
\begin{figure}[!htb]
    \centering
    \begin{tikzpicture}[scale=0.8, every node/.style={scale=1.0}]
    \begin{axis}[
        width  = 1.0*\textwidth,
        height = 9.5cm,
        ymin=0.0,ymax=1.2,
        ytick={0,0.2,0.4,0.6,0.8,1.0},
        major x tick style = transparent,
        ybar=5*\pgflinewidth,
        bar width=17.0pt,
%        ymajorgrids = true,
        xlabel = {Prediction type},
        ylabel = {Accuracy},
        symbolic x coords={Passing pedestrian,Passing biker,Crossing pedestrian,Crossing biker,Overall},
        xticklabels={
        		$\mbox{Passing}\atop\mbox{pedestrian}$,
        		$\mbox{Passing}\atop\mbox{biker}$,
        		$\mbox{Crossing}\atop\mbox{pedestrian}$,
        		$\mbox{Crossing}\atop\mbox{biker}$,
        		$\mbox{Overall}\atop\mbox{accuracy}$},
        xtick=data,
	y tick label style={
    		/pgf/number format/.cd,
   		fixed,
   		fixed zerofill,
    		precision=2},
%	yticklabel pos=right,
%        xtick = data,
        x tick label style={
		font=\small,
%		anchor=north east,
%		inner sep=0mm
		},
%		font=\small},
%        scaled y ticks = false,
	%%%%% numbers on bars and rotated
        nodes near coords,
        every node near coord/.append style={rotate=90, 
        								   anchor=west,
								   font=\footnotesize,
								   /pgf/number format/.cd,
								   	fixed zerofill,
									precision=4
								   },
        %%%%%
%        enlarge x limits=0.03,
        enlarge x limits=0.125,
        legend cell align=left,
        legend pos=south east,
%        legend style={
%                at={(1,1.05)},
%                anchor=south east,
%	        nodes={rotate=90},%%%%% rotate text in legend
%                at={(0.125,0)},
%                at={(0.125,0)},
%                at={(0.8775,0)},
%                at={(0.89,0.02)},
%                anchor=south,
%                column sep=1ex
%        },
%        axis x line*=bottom
    ]
\addplot[fill=blue,opacity=1.00]
coordinates {
(Passing pedestrian,0.8125)
(Passing biker,1.0000)
(Crossing pedestrian,0.9367)
(Crossing biker,0.7500)
(Overall,0.9035)
};
\addplot[fill=red,opacity=1.00]
coordinates {
(Passing pedestrian,0.7975)
(Passing biker,0.8571)
(Crossing pedestrian,0.8984)
(Crossing biker,0.6667)
(Overall,0.8472)
};
\addplot[fill=green,opacity=1.00]
coordinates {
(Passing pedestrian,0.7059)
(Passing biker,0.9333)
(Crossing pedestrian,0.9143)
(Crossing biker,0.6250)
(Overall,0.7982)
};
\legend{Summer school,Shopping mall,Park}
\end{axis}
\end{tikzpicture}
    \caption{Accuracies for prediction types} 
    \label{fig:location_accuracies}
\end{figure}

%%%%% I don't understand these numbers. Need to make this clearer ?????
%%%%% Should also provide a bar graph of the bottom-line results
%%%%% Maybe explain what these numbers mean in terms of the crosswalk lights,
%%%%% that is, how often will the lights be activated when nobody is present
%%%%% and how often will the lights fail to activate when somebody is crossing...
Although the accuracy fluctuated with the location and the number of people at the location, 
the overall accuracy of correctly predicting pedestrian behavior across all locations 
was~0.8496. The overall accuracy of predicting pedestrians (or bikers) without intentions to cross
the road was~0.7943 whereas the overall accuracy of predicting pedestrians (or bikers) with intentions
to cross the road was~0.8845. These results suggest that if a pedestrian or biker were to try to cross a 
street similar to the ones in the study, there will be a~0.8845 chance our program will recognize and trigger
the crosswalk signal before they leave, helping an average of~81.67 pedestrians per hour in our study.
If a pedestrian or biker were to pass by the camera without intentions to cross the street, 
there is a~0.7943 chance the program ignores them and a~0.2057 chance the program falsely activates the crosswalk, 
occurring an average of~12 times per hour in our experiment. Figure \ref{fig:location_accuracies} shows 
more detailed statistics regarding the prediction accuracies.

Furthermore, the average false alarm rate for all locations was~0.00037 per prediction.
Since our model currently makes predictions with a~1 second delay for a total of~86,400 predictions per day, 
32 of these predictions are expected to be false alarms, resulting in the crosswalk signal being
activated when no pedestrians are nearby.

\subsection{Discussion}

The results above indicate that our CNN-based system has the ability to detect 
pedestrian and biker intentions at various real-world locations. The system has a relatively 
high accuracy rate of~0.8496 and a low false alarm rate of~0.00037. Thus, a deployment of the model 
in a real world environment would generally help pedestrians who fail to activate crosswalk lights,
without disrupting the traffic flow too often.

As seen in Figure \ref{fig:location_accuracies}, the summer school had the highest overall accuracy of~0.9035 and 
yielded the highest accuracy for all prediction types. The shopping mall had the second highest accuracy~0.8472, 
although sometimes had lower accuracy than the park, which had an overall accuracy of~0.7982.
Furthermore, the accuracy trends for the different prediction types were generally the same across the three locations, 
where bikers not crossing the street were predicted most accurately and bikers crossing the street were predicted
least accurately.

Our data from Figure \ref{fig:location_accuracies} indicates the method of transportation greatly influences the
accuracy of predictions. Across all three locations, the bikers were predicted more accurately than pedestrians 
by up to~18.75\% when both did not have intentions to cross the street, whereas pedestrians were predicted more
accurately than bikers by up to~11.61\% when both had intentions to cross the street. The contrast in accuracy is 
likely due to differences between biking and walking, such as speed, that work better with various thresholds 
but worse in others. Bikers are less likely to be detected for multiple images since they are 
faster, which increases the accuracy when they are passing by but decreases the accuracy when they cross the street. 
Although our threshold selection was optimized to produce the highest overall accuracy, distinct thresholds
might be optimal for bikers and pedestrians

Another result from this experiment is how individual pedestrian behavior influences the accuracy of predictions. 
In the shopping mall location, the program accurately predicted a passing pedestrian with a~0.7975 
accuracy, which is much higher than the~0.7059 accuracy from the park crosswalk. For pedestrians 
crossing the street, however, the shopping mall had a lower accuracy of~0.8984 
compared to the park accuracy of~0.9143. The summer school's accuracy remained sightly higher than that of
the shopping mall and park. After reviewing the footage from each location, we believe this difference 
in accuracy can be attributed to the volume and behavior of people at each location. For instance, 
pedestrians taking a stroll around the park usually walked slower than the shopper and students,
which decreases the park's accuracy when pedestrians pass by the camera slowly but increases the accuracy when
they cross the road slowly. Other details unique to each location include the summer school students being more 
cautious when crossing the road and the shopping mall pedestrians having to stand in front of the camera longer
before crossing the street to wait for the crosswalk signal. These differences present distinct challenges that
affect the accuracy of each location, and our results could likely be improved significantly, if we optimized the
setting for an individual location.

The difference in location also resulted in different false alarm results.
For instance, the park location resulted in~3 false alarms,~2 more than the shopping mall and~3 more than the summer school. 
Although these differences seem minor, they can result in a difference of dozens of false alarms per day. 
We believe this difference is due to an inadequate amount of data that feature a park 
as its background. As a result, the CNN model has more trouble differentiating between 
pedestrians and streets, so small changes in the view produce a false alarm.
The opposite seems to be true for the summer school location, which produced~0 false alarms.
This is most likely due to our dataset featuring relatively more images of streets similar to the ones around
a school. For the CNN model to produce less false alarms for different locations, it is therefore 
important to have adequate training data with a similar background to the deployed location.

\section{Conclusion and Future Works}\label{chap:future}

To improve the safety of crossing the street, we developed a machine learning approach 
that could effectively differentiate between the street, pedestrians, and bike riders. 
After collecting~2000 images for our dataset, we used the GoogLeNet CNN architecture to train a model,
with a grid search employed to tune the hyperparameters. We then created an experimental system 
that detects pedestrians in real time. We tested the system in three real-world settings and the results
indicate that the system is practical.

The dataset used to train our CNN model could be improved in various ways.
Our dataset is relatively small, and a larger dataset might improve the accuracy.
The evidence from our park crosswalk experiments indicate that
images that are specific to a particular location will likely significantly increase the accuracy. 
External factors, such as the weather or light levels might have
an effect on accuracy. A crosswalk safety system may be even more important for more extreme weather conditions, 
so it could be useful to gather ample data to account for such situations.

Our selected threshold of~$(n,t)=(5,3)$ was set to maximize the overall accuracy for bike riders and pedestrians.
However, our experimental results indicate that the individual accuracies may be improved if
different values of~$n$ are used to identify bikers and pedestrians. As a result, we believe it
will be useful to consider a detector that combines distinct systems for different methods of transportation.
Such an approach can account for minor differences, such as speed, that might be currently limiting our accuracy.

Based on our experiment results, the system performs somewhat less well when predicting the behavior of 
pedestrians and bikers with no intentions to cross the street. Since our underlying CNN models is
focused primarily on object classification, it may be useful to include other machine learning models. 
For example, models are designed to track an object, rather than classify it, could be particularly useful
in for an automated crosswalk signaling system. In particular, the YOLO algorithm is a fast object detection 
algorithm that locates the position of an object in an image and can track its movement \cite{yolo}. 

\bibliographystyle{plain}

\bibliography{references.bib}

\section*{Appendix}

In this appendix, we display the GoogLeNet architecture that is used in our program. In our research, we custom-train our model instead of using the pre-trained model. This allows us to have custom input size and more control over the model.

%%%%% This is too small to read. Can it be rendered in a serpentine manner, so that things can be
%%%%% enlarged, but still fit on one page ?????
\begin{figure}[!htb]
    \centering
%    \vspace{-3.21cm}
    \includegraphics[width=0.53\textwidth]{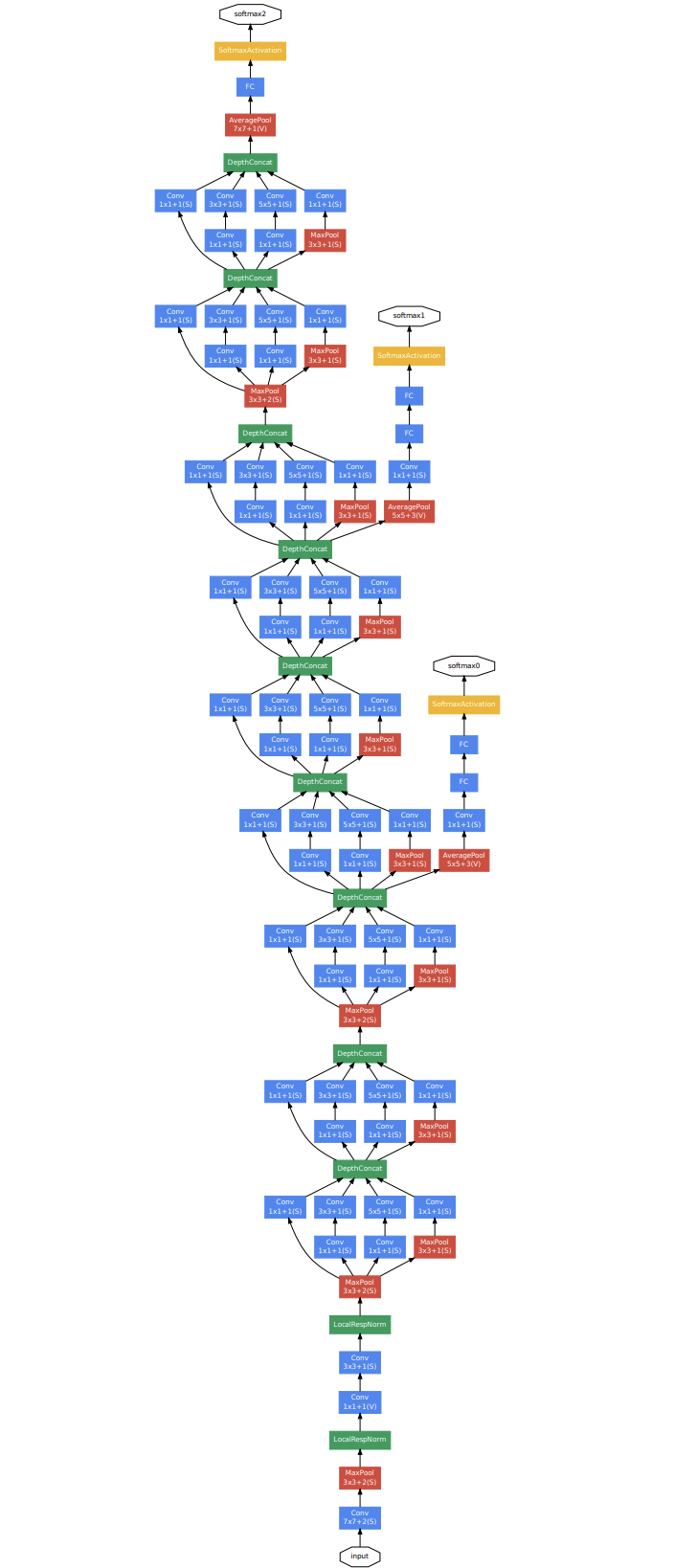}
    \caption{The GoogLeNet architecture~\cite{convNet}} 
    \label{fig:googlenet}
    \vspace{-61pt}
\end{figure}

\end{document}